%% file: SentRepDiscourse.tex
\documentclass[11pt,a4paper]{article}
\usepackage[nohyperref]{emnlp2017}
\usepackage{times}
\usepackage{latexsym}
\usepackage{booktabs}
\usepackage{multirow}
\usepackage{graphicx}
\usepackage[hidelinks,draft]{hyperref}
\usepackage[]{color}
\usepackage{appendix}

\emnlpfinalcopy

\def\emnlppaperid{***}

\title{Discourse-Based Objectives\\ for Fast Unsupervised Sentence Representation Learning}

\author{
Yacine Jernite\\Department of Computer Science\\New York University\\\texttt{yacine.jernite@nyu.edu}
         \And  
Samuel R. Bowman\\ Department of Linguistics\\and Center for Data Science\\New York University\\\texttt{bowman@nyu.edu}
         \And  
David Sontag\\ Department of EECS\\Massachussets Institute\\of Technology\\\texttt{dsontag@mit.edu}
}

\date{}

\begin{document}

\maketitle
\input{abstract}
\input{introduction}
\input{related}

\input{objectives}

\input{results}

\input{conclusion}



\bibliography{SentRepDiscourse}
\bibliographystyle{emnlp_natbib}

\clearpage

\appendix
\input{supplementary}

\end{document}

%% file: abstract.tex
\begin{abstract}
This work presents a novel objective function for the unsupervised training of neural network sentence encoders. It exploits signals from paragraph-level discourse coherence  to train these models to understand text. Our objective is purely discriminative, allowing us to train models many times faster than was possible under prior methods, and it yields models which perform well in extrinsic evaluations. 
\end{abstract}

%% file: introduction.tex
\section{Introduction}

Modern artificial neural network approaches to  natural language understanding tasks like translation \citep{sutskever2014sequence,DBLP:conf/ssst/ChoMBB14},  summarization \citep{DBLP:conf/emnlp/RushCW15}, and classification \citep{DBLP:conf/naacl/YangYDHSH16} depend crucially on subsystems called \textit{sentence encoders} that construct distributed representations for sentences. These encoders are typically implemented as convolutional \citep{DBLP:conf/emnlp/Kim14}, recursive \citep{socher2013recursive}, or recurrent neural networks \citep{DBLP:conf/interspeech/MikolovKBCK10} operating over a sentence's words or characters \citep{DBLP:conf/nips/ZhangZL15,DBLP:conf/aaai/KimJSR16}. 

Most of the early successes with sentence encoder-based models have been on tasks with ample training data, where it has been possible to train the encoders in a fully-supervised end-to-end setting. However, recent work  has shown some success in using unsupervised pretraining with unlabeled data to both improve the performance of these methods and extend them to lower-resource settings \citep{dai2015semi, DBLP:conf/nips/KirosZSZUTF15,DBLP:journals/corr/BajgarKK16}. 

\begin{figure}[t]
  \hspace{-0.5in}
  \begin{center}
  \includegraphics[width=0.45\textwidth]{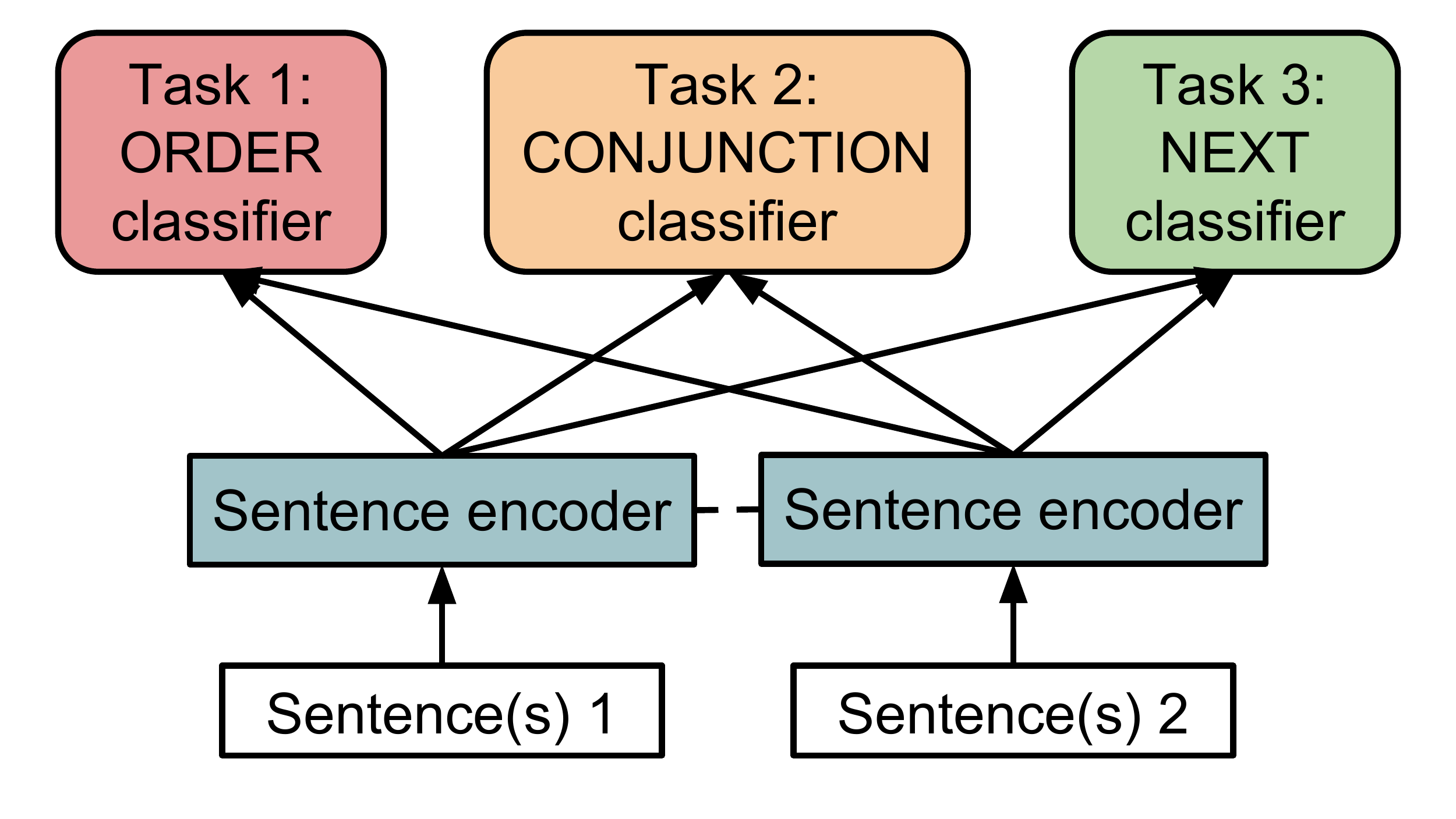}
  \end{center}
   \vspace{-0.1in}
  \caption{\label{fig:joint_archi} We train a sentence encoder (shown as two copies with shared parameters) on three discourse-based objectives over unlabeled text.
  } 
\end{figure}

This paper presents a set of methods for unsupervised pretraining that train sentence encoders to recognize \emph{discourse coherence}. When reading text, human readers have an expectation of coherence from one sentence to the next. In most cases, for example, each  sentence in a text should be both interpretable in context and relevant to the topic under discussion. Both of these properties depend on an understanding of the local context, which includes both relatively knowledge about the state of the world and the specific meanings of previous sentences in the text. Thus, a model that is successfully trained to recognize discourse coherence must be able to understand the meanings of sentences as well as relate them to key pieces of knowledge about the world.

\citet{DBLP:journals/cogsci/Hobbs79} presents a formal treatment of this phenomenon. He argues that for a discourse (here, a text) to be interpreted as coherent, any two adjacent sentences must be related by one of a few set kinds of \emph{coherence relations}. For example, a sentence might be followed by another that elaborates on it, parallels it, or contrasts with it. While this treatment may not be adequate to cover the full complexity of language understanding, it allows Hobbs to show how identifying such relations depends upon sentence understanding, coreference resolution, and commonsense reasoning.

Recently proposed techniques \citep{DBLP:conf/nips/KirosZSZUTF15,ramachandran2016unsupervised} succeed in exploiting discourse coherence information of this kind to train sentence encoders, but rely on generative objectives which require models to compute the likelihood of each word in a sentence at training time. In this setting, a single epoch of training on a typical (76M sentence) text corpus can take weeks, making further research difficult, and making it nearly impossible to scale these methods to the full volume of available unlabeled English text. In this work, we propose alternative objectives which exploit much of the same  coherence information at greatly reduced cost. 

In particular, we propose three fast coherence-based pretraining tasks, show that they can be used together effectively in multitask training (Figure~\ref{fig:joint_archi}), and evaluate models trained in this setting on the training tasks themselves and on standard text classification tasks.\footnote{All code, resources, and models involved in these experiments will be made available upon publication.} We find that our approach makes it possible to learn to extract broadly useful sentence representations in hours.
 

%% file: related.tex
\section{Related Work}
\label{sec:related}

This work is inspired most directly by the Skip Thought approach of \citet{DBLP:conf/nips/KirosZSZUTF15}, which introduces the use of paragraph-level discourse information for the unsupervised pretraining of sentence encoders.  Since that work, three other papers have presented improvements to this method (the SDAE of \citealt{DBLP:conf/naacl/HillCK16}, also \citealt{gan2016unsupervised,ramachandran2016unsupervised}). These improved methods are based on techniques and goals that are similar to ours, but all three involve models that explicitly generate full sentences during training time at considerable computational cost. 

In closely related work, \citet{logeswaran2016sentence} present a model that learns to order the sentences of a paragraph. While they focus on learning to assess coherence, they show positive results on measuring sentence similarity using their trained encoder. Alternately, the FastSent model of \citet{DBLP:conf/naacl/HillCK16} is designed to work dramatically more quickly than systems like Skip Thought, but in service of this goal the standard sentence encoder RNN is replaced with a low-capacity CBOW model. Their method does well on existing semantic textual similarity benchmarks, but its insensitivity to order places an upper bound on its performance in more intensive extrinsic language understanding tasks.

Looking beyond work on unsupervised pretraining:
\citet{DBLP:conf/emnlp/LiH14a} and \citet{li2016neural} use representation learning systems to directly model the problem of sentence order recovery, but focus primarily on intrinsic evaluation rather than transfer. 
\citet{DBLP:conf/acl/WangC16} train sentence representations for use as context in language modeling. In addition, \citet{DBLP:journals/corr/JiHE16} treat discourse relations between sentences as latent variables and show that this yields improvements in language modeling in an extension of the document-context model of \citet{DBLP:journals/corr/JiCKDE15}.

Outside the context of representation learning, there has been a good deal of work in NLP on discourse coherence, and on the particular tasks of sentence ordering and coherence scoring. \citet{barzilay2008modeling} provide thorough coverage of this work.

%% file: objectives.tex
\section{Discourse Inspired Objectives}
\label{sec:objectives}

\begin{table}[t]
\begin{center}\small
\begin{tabular}{lll}
\toprule
\bf Sentence Pair              & \bf Label & \bf Relation \\
\midrule
\it A strong one at that.  & \multirow{2}{*}{\sc Y} & \multirow{2}{*}{elaboration} \\
\it Then I became a woman. &   &  \\
\midrule
\it I saw flowers on the ground.         & \multirow{2}{*}{\sc N}  & \multirow{2}{*}{list} \\
\it I heard birds in the trees.           &   &  \\
\midrule
\it It limped closer at a slow pace.& \multirow{2}{*}{\sc N} & \multirow{2}{*}{spatial} \\
\it Soon it stopped in front of us. & &  \\
\midrule
\it I kill Ben, you leave by yourself.& \multirow{2}{*}{\sc Y} & \multirow{2}{*}{time} \\
\it I kill your uncle, you join Ben.  &  &  \\
\bottomrule
\end{tabular}
\end{center}
\caption{The binary \textsc{order} objective. Discourse relation labels are provided for the reader, but are not available to the model.}
\label{tab:order-task}
\end{table}

In this work, we propose three objective functions for use over paragraphs extracted from unlabeled text. Each captures a different aspect of discourse coherence and together the three can be used to train a single encoder  to extract broadly useful sentence representations.

\paragraph{Binary Ordering of Sentences} Many coherence relations have an inherent direction. For example, if $S_1$ is an elaboration of $S_0$, $S_0$ is not generally an elaboration of $S_1$. Thus, being able to identify these coherence relations implies an ability to recover the original order of the sentences. Our first task, which we call {\sc order}, consists in taking pairs of adjacent sentences from text data, switching their order with probability 0.5, and training a model to decide whether they have been switched. Table \ref{tab:order-task} provides some examples of this task, along with the kind of coherence relation that we assume to be involved. It should be noted that since some of these relations are unordered, it is not always possible to recover the original order based on discourse coherence alone (see \emph{e.g.} the \emph{flowers} / \emph{birds} example).

\begin{table}[t]
\begin{center}\small
\begin{tabular}{l}
\toprule
\bf Context \\
\midrule
\it No, not really.\\
\it I had some ideas, some plans.\\
\it But I never even caught sight of them.\\
\midrule
\bf Candidate Successors \\
\midrule
1. \it There's nothing I can do that compares that. \\
2. \it Then one day Mister Edwards saw me. \\
\textbf{3. \textit{I drank and that was about all I did.}} \\
4. \it And anyway, God's getting his revenge now. \\
5. \it He offered me a job and somewhere to sleep. \\
\bottomrule
\end{tabular}
\end{center}
\caption{The \textsc{next} objective.}
\label{tab:next-task}
\end{table}

\paragraph{Next Sentence} Many coherence relations are transitive by nature, so that any two sentences from the same paragraph will exhibit some coherence. However, two adjacent sentences will generally be more coherent than two more distant ones. This leads us to formulate the {\sc next} task: given the first three sentences of a paragraph and a set of five candidate sentences from later in the paragraph, the model must decide which candidate immediately follows the initial three in the source text. Table~\ref{tab:next-task} presents an example of such a task: candidates 2 and 3 are coherent with the third sentence of the paragraph, but the elaboration (3) takes precedence over the progression (2).

\begin{table}[t]
\begin{center}\small
\begin{tabular}{ll}
\toprule
\bf Sentence Pair & \bf Label \\
\midrule
\it He had a point.               & \sc return  \\
\it For good measure, I pouted.   & \it (Still) \\
\midrule
\it It doesn't hurt at all.       & \sc strengthen \\
\it It's exhilarating.           & \it (In fact)  \\
\midrule
\it The waterwheel hammered on.   & \sc contrast  \\
\it There was silence.            & \it (Otherwise) \\
\bottomrule
\end{tabular}
\end{center}
\caption{The \textsc{conjunction} objective. Discourse relation labels are shown with the text from which they were derived.}
\label{tab:conj-task}
\end{table}

\paragraph{Conjunction Prediction} Finally, information about the coherence relation between two sentences is sometimes apparent in the text \citep{miltsakaki2004penn}: this is the case whenever the second sentence starts with a conjunction phrase. To form the {\sc conjunction} objective, we create a list of conjunction phrases and group them into nine categories (see supplementary material). We then extract from our source text all pairs of sentences where the second starts with one of the listed conjunctions, give the system the pair without the phrase, and train it to recover the conjunction category. Table \ref{tab:conj-task} provides examples.

%% file: results.tex
\section{Experiments}
\label{sec:results}

In this section, we introduce our training data and methods, present qualitative results and comparisons among our three objectives, and close with quantitative comparisons with related work.

\paragraph{Experimental Setup}  We train our models on a combination of data from BookCorpus  \citep{DBLP:conf/iccv/ZhuKZSUTF15}, the Gutenberg project \citep{DBLP:journals/crossroads/Stroube03}, and Wikipedia. After sentence and word tokenization \citep[with NLTK;][]{DBLP:conf/acl/Bird06} and lower-casing, we identify all paragraphs longer than 8 sentences and extract a \textsc{next} example from each, as well as pairs of sentences for the \textsc{order} and \textsc{conjunction} tasks. This gives us 40M  examples for \textsc{order}, 1.4M for \textsc{conjunction}, and 4.1M for \textsc{next}.

Despite having recently become a standard dataset for unsupervised learning, BookCorpus does not exhibit sufficiently rich discourse structure to allow our model to fully succeed---in particular, some of the conjunction categories are severely under-represented. Because of this,  we choose to train our models on text from all three sources. While this precludes a strict apples-to-apples comparison with other published results, 
our goal in extrinsic evaluation is simply to show that our method makes it possible to learn useful representations quickly, rather than to demonstrate the superiority of our learning technique given fixed data and unlimited time.

We consider three sentence encoding models: a simple 1024D sum-of-Words (CBOW) encoding, a 1024D GRU  recurrent neural network \citep{DBLP:conf/ssst/ChoMBB14}, and a 512D bidirectional GRU RNN (BiGRU). All three use FastText \citep{DBLP:journals/corr/JoulinGBM16} pre-trained word embeddings\footnote{https://github.com/facebookresearch/fastText/ blob/master/pretrained-vectors.md} to which we apply a Highway transformation \citep{DBLP:journals/corr/SrivastavaGS15}. 
The encoders are trained jointly with three bilinear classifiers for the three objectives (for the {\sc{next}} examples, the three context sentences are encoded separately and their representations are concatenated). We perform stochastic gradient descent with AdaGrad \citep{DBLP:journals/jmlr/DuchiHS11}, subsampling \textsc{conjunction} and \textsc{next} by a factor of 4 and 6 respectively (chosen using held-out accuracy averaged over all three tasks on held out data after training on 1M examples). In this setting, the BiGRU model takes 8 hours to see all of the examples from the BookCorpus dataset at least once. For ease of comparison, we train all three models for exactly 8 hours.

\paragraph{Intrinsic and Qualitative Evaluation} Table~\ref{tab:joint} compares the performance of different training regimes along two axes: encoder architecture and whether we train one model per task  or one joint model. As expected, the more complex bidirectional GRU architecture is required to capture the appropriate sentence properties, although CBOW still manages to beat the simple GRU (the slowest model), likely by virtue of its substantially faster speed, and correspondingly greater number of training epochs. Joint training does appear to be effective, as both the {\sc order} and {\sc next} tasks benefit from the information provided by {\sc conjunction}. Early experiments on the external evaluation also show that the joint BiGRU model substantially outperforms each single model.

\begin{table}[t!]
\small
\begin{centering}
\begin{tabular}{l r r r}
\toprule
 & \sc conjunction & \sc order & \sc next  \\
\midrule
CBOW joint   & 42.8 & 56.6 & 27.7 \\  
GRU joint   & 39.5 & 54.3 & 25.9 \\  
BiGRU joint & 45.1 & \bf 58.3 & \bf 30.2 \\
\midrule
BiGRU single & \bf 45.5 & 57.1 & 29.3 \\
\bottomrule
\end{tabular}
\end{centering}
\caption{Intrinsic evaluation results.}
\label{tab:joint}
\end{table}

Table \ref{tab:neighbors} and the supplementary material show nearest neighbors in the trained BiGRU's representation space for a random set of seed sentences. 
We select neighbors from among 400k held-out sentences. The encoder appears to be especially sensitive to high-level syntactic structure.

\begin{table}[t!]
\small
\begin{centering}
\begin{tabular}{l}
\toprule
\textbf{\textit{Grant laughed and complied with the suggestion.}} \\
\midrule
{\emph{Pauline stood for a moment in complete bewilderment.}} \\
{\emph{Her eyes narrowed on him, considering.}} \\
{\emph{Helena felt her face turn red hot.}} \\
{\emph{Her face remained expressionless as dough.}} \\
\bottomrule
\end{tabular}
\end{centering}
\caption{The nearest neighbors for a sentence.}
\label{tab:neighbors}
\end{table}

\paragraph{Extrinsic Evaluation} We evaluate the quality of the encoder learned by our system, which we call DiscSent, by using the sentence representations it produces in a variety of sentence classification tasks. We follow the settings of \citet{DBLP:conf/nips/KirosZSZUTF15} on paraphrase detection \citep[MSRP;][]{DBLP:conf/coling/DolanQB04}, subjectivity evaluation \citep[SUBJ;][]{DBLP:conf/acl/PangL04} and question classification \citep[TREC;][]{DBLP:conf/trec/Voorhees01a}. 

\begin{table}[t!]
\small
\begin{centering}
\begin{tabular}{lrrrr}
\toprule
\bf Model         & \bf Time & \bf MSRP & \bf TREC & \bf SUBJ \\
\toprule
FastSent$^1$      & \multirow{2}{*}{$\approx$13h}& 72.2 & 76.8 & 88.7 \\ 
FastSent+AE$^1$   &                       & 71.2 & 80.4 & 88.8 \\ 
\midrule
SDAE$^1$          & \multirow{2}{*}{192h} & \bf 76.4 & 77.6 & 89.3 \\ 
SDAE+embed$^1$    &                       & 73.7 & 78.4 & 90.8 \\ 
\midrule
SkipT biGRU$^2$   & \multirow{3}{*}{336h} & 71.2 & 89.4 & 92.5 \\ 
SkipT GRU$^2$     &                       & 73.0 & 91.4 & 92.1 \\ 
SkipT+feats$^2$   &                       & 75.8 & \bf 92.2 & \bf 93.6 \\ 
\midrule
Ordering model$^3$& \multirow{3}{*}{48h}  & 72.3 & -- & -- \\
Ordering+embed$^3$  &                     & 74.0 & -- & -- \\
+embed+SkipT$^3$  &                       & 74.9 & -- & -- \\
\midrule
DiscSent biGRU    & \multirow{3}{*}{\bf 8h} & 71.6 & 81.0 & 88.6 \\
DiscSent+unigram  &                         & 72.5 & 87.9 & 92.7 \\
DiscSent+embed    &                         & 75.0 & 87.2 & 93.0 \\
\bottomrule
\end{tabular}
\end{centering}
\caption{Text classification results, including training time. +embed lines combine the sentence encoder output with the sum of the pretrained word embeddings for the sentence. +unigram lines do so using embeddings learned for each target task without pretraining. 
+feats varies by task. References: 
$^1$\citet{DBLP:conf/naacl/HillCK16}
$^2$\citet{DBLP:conf/nips/KirosZSZUTF15}
$^3$\citet{logeswaran2016sentence}}
\label{tab:external}
\end{table}

Overall, our system performs comparably with the SDAE and  Skip Thought 
approaches
with a drastically shorter training time. Our system  also compares favorably to the similar discourse-inspired method of \citet{logeswaran2016sentence}, achieving similar results on MSRP in a sixth of their training time.

%% file: conclusion.tex
\section{Conclusion}
\label{sec:conclusion}

In this work, we introduce three new training objectives for unsupervised sentence representation learning inspired by the notion of discourse coherence, and use them to train a sentence representation system in competitive time, from 6 to over 40 times shorter than comparable methods, while obtaining comparable results on external evaluations tasks. We hope that the tasks that we introduce in this paper will prompt further research into discourse understanding with neural networks, as well as into strategies for unsupervised learning that will make it possible to use unlabeled data to train and refine a broader range of models for language understanding tasks.

%% file: supplementary.tex
\onecolumn
\section*{Supplement to:\\ Discourse-Based Objectives for Fast Unsupervised Sentence Representation Learning\\}

Table~\ref{tab:conjunction} lists the conjunction phrases and groupings used.
Table~\ref{tab:neighbors_large} (next page) shows the Euclidean nearest neighbors of a sample of sentences in our representation space.


\begin{table}[h]
\centering\small
\begin{tabular}{ccccc}
\toprule
\multicolumn{2}{c}{\textbf{addition}} & \multicolumn{2}{c}{\textbf{contrast}}  & \textbf{time} \\
again   & furthermore & anyway  & contrarily  & meanwhile  \\
also    & moreover    & however & conversely  & next       \\
besides & in addition & instead & nonetheless & then       \\
finally & & nevertheless        & in contrast & now        \\
further & & otherwise           & rather      & thereafter \\
\midrule
\textbf{result}  & \textbf{specific}  & \textbf{compare} & \textbf{strengthen} & \textbf{return}    \\
accordingly  & namely       & likewise  & indeed & still \\
consequently & specifically & similarly & in fact &      \\
\cline{5-5}
hence        & notably      & & & \textbf{recognize}     \\
thus         & that is      & & & undoubtedly            \\
therefore    & for example  & & & certainly              \\
\bottomrule
\end{tabular}
\caption{Grouping of conjunctions.}
\label{tab:conjunction}
\end{table}

\clearpage


\begin{table}[h!]
\small
\begin{centering}
\begin{tabular}{l}
{\bf His main influences are Al Di, Jimi Hendrix, Tony, JJ Cale, Malmsteen and Paul Gilbert.} \\
\midrule
The album features guest appearances from Kendrick Lamar, Schoolboy Q, 2 Chainz, Drake, Big. \\
The production had original live rock, blues, jazz, punk, and music  composed and arranged by Steve and Diane Gioia. \\
There are 6 real drivers in the game: Gilles, Richard Burns, Carlos Sainz, Philippe, Piero, and Tommi. \\
Other rappers that did include Young Jeezy, Lil Wayne, Freddie Gibbs, Emilio Rojas, German rapper and Romeo Miller. \\
\\
{\bf Grant laughed and complied with the suggestion.} \\
\midrule
Pauline stood for a moment in complete bewilderment. \\
Her eyes narrowed on him, considering. \\
Helena felt her face turn red hot. \\
Her face remained expressionless as dough. \\
\\
{\bf Items can be selected specifically to represent characteristics that are not as well represented in natural language.} \\
\midrule
Cache manifests can also use relative paths or even absolute urls as shown below. \\
Locales can be used to translate into different languages, or variations of text, which are replaced by reference. \\
Nouns can only be inflected for the possessive, in which case a prefix is added. \\
Ratios are commonly used to compare banks, because most assets and liabilities of banks are constantly valued at market values. \\
\\
{\bf A group of generals thus created a secret organization, the united officers' group, in order to oust Castillo from power.} \\
\midrule
The home in Massachusetts is controlled by a private society organized for the purpose, with a board of fifteen trustees in charge. \\
A group of ten trusted servants men from the family were assigned to search the eastern area of the island in the area. \\
The city is divided into 144 administrative wards that are grouped into 15 boroughs. each of these wards elects a councillor. \\
From 1993 to 1994 she served as US ambassador to the United Nations commission on the status of women. \\
\\
{\bf As a result of this performance, Morelli's play had become a polarizing issue amongst Nittany Lion fans.} \\
\midrule
In the end, Molly was deemed to have more potential, eliminating Jaclyn despite having a stellar portfolio. \\
As a result of the Elway connection, Erickson spent time that year learning about the offense with Jack. \\
As a result of the severe response of the czarist authorities to this insurrection, had to leave Poland. \\
Another unwelcome note is struck by the needlessly aggressive board on the museum which has already been mentioned. \\
\\
{\bf Zayd Ibn reported , ``we used to record the Quran from parchments in the presence of the messenger of god.''} \\
\midrule
Daniel Pipes says that ``primarily through ``the protocols of the Elders of Zion'', the whites spread these charges to [\ldots]'' \\ 
Sam wrote in ``'' (1971) that Howard's fiction was ``a kind of wild West in the lands of unbridled fantasy.'' \\
said , the chancellor ``elaborately fought for an European solution'' in the refugee crisis, but this was ``out of sight''. \\
Robert , writing for ``The New York Post'', states that, ``in Mellie , the show has its most character [\ldots]'' \\ 
\\
{\bf Many ``Crimean Goths'' were Greek speakers and many Byzantine citizens were settled in the region called [\ldots]} \\ 
\midrule
The personal name of ``Andes'', popular among the Illyrians of southern Pannonia and much of Northern Dalmatia [\ldots] \\ 
is identified by the Chicano as the first settlement of the people in North America before their Southern migration [\ldots] \\ 
The range of ``H.'' stretches across the Northern and Western North America as well as across Europe [\ldots] \\ 
The name ``Dauphin river'' actually refers to two closely tied communities; bay and some members of Dauphin river first nation.\\
\\
{\bf She smiled and he smiled in return.} \\
\midrule
He shook his head and smiled broadly. \\
He laughed and shook his head. \\
He gazed at her in amazement. \\
She sighed and shook her head at her foolishness. \\
\\
{\bf The jury returned a verdict of “not in the Floyd cox case, in which he was released immediately.} \\
\midrule
The match lasted only 1 minute and 5 seconds, and was the second quickest bout of the division. \\
His results qualified him for the Grand Prix final, in which he placed 6th overall. \\
The judge stated that the prosecution had until march 1, 2012, to file charges. \\
In November, he reached the final of the Ruhr Open, but lost 4\u20130 against Murphy. \\
\\
{\bf Here was at least a slight reprieve.} \\
\midrule
The monsters seemed to be frozen in time. \\
This had an impact on him. \\
That was all the sign he needed. \\
So this was disturbing as hell. \\
\end{tabular}
\end{centering}
\caption{Nearest neighbor examples}
\label{tab:neighbors_large}
\end{table}